\documentclass[letterpaper]{article} % DO NOT CHANGE THIS
\usepackage{aaai23}  
\usepackage{times}  % DO NOT CHANGE THIS
\usepackage{helvet}  % DO NOT CHANGE THIS
\usepackage{courier}  % DO NOT CHANGE THIS
\usepackage[hyphens]{url}  % DO NOT CHANGE THIS
\usepackage{graphicx} % DO NOT CHANGE THIS
\usepackage{natbib}  % DO NOT CHANGE THIS AND DO NOT ADD ANY OPTIONS TO IT
\usepackage{caption} % DO NOT CHANGE THIS AND DO NOT ADD ANY OPTIONS TO IT
\usepackage{algorithm}
\usepackage{algorithmic}
\usepackage{amssymb}
\usepackage{amsmath,amsfonts}
\usepackage{multirow}
\usepackage{booktabs}

\usepackage{url}
\usepackage{graphicx} %插入图片的宏包
\usepackage{float} %设置图片浮动位置的宏包
\usepackage{booktabs} % 表格加粗
\usepackage{makecell} % 表格内元素居中
\usepackage{multirow} % 表格多行时需要
\usepackage{amsfonts} % 导入希腊字体
\usepackage{bm}  % 给公式加粗
\usepackage{bbding}
\usepackage{color,xcolor}

\usepackage{newfloat}
\usepackage{listings}
\DeclareCaptionStyle{ruled}{labelfont=normalfont,labelsep=colon,strut=off} % DO NOT CHANGE THIS
\floatstyle{ruled}
\newfloat{listing}{tb}{lst}{}
\floatname{listing}{Listing}
\begin{document}

\title{PPN: Parallel Pointer-based Network for Key Information Extraction with Complex Layouts}
\author{
    Kaiwen Wei\textsuperscript{\rm 1},
    Jie Yao\textsuperscript{\rm 2},
    Jingyuan Zhang\textsuperscript{\rm 1}, 
    Yangyang Kang\textsuperscript{\rm 1}\thanks{Corresponding Author.}, 
    Fubang Zhao\textsuperscript{\rm 1}, \\
    Yating Zhang\textsuperscript{\rm 1},
    Changlong Sun\textsuperscript{\rm 1},
    Xin Jin\textsuperscript{\rm 2},
    Xin Zhang\textsuperscript{\rm 2}
}
\affiliations{
    %Afiliations
    \textsuperscript{\rm 1}DAMO Academy, Alibaba Group \\
    \textsuperscript{\rm 2}Ant Group \\
    \{zhuping.wkw, yangyang.kangyy\}@alibaba-inc.com
    
    % \textsuperscript{\rm 2}Key Laboratory of Network Information System Technology (NIST), Aerospace Information Research Institute
   
    % \textsuperscript{\rm 3}School of Electronic, Electrical and Communication Engineering, University of Chinese Academy of Sciences\\
    
   % zhanglinhao20@mails.ucas.ac.cn, jinlimails@gmail.com\\
}

\maketitle

\begin{abstract}
\begin{quote}
Key Information Extraction (KIE) is a challenging multimodal task that aims to extract structured value  semantic entities from visually rich documents. Although significant progress has been made, there are still two major challenges that need to be addressed. Firstly, the layout of existing datasets is relatively fixed and limited in the number of semantic entity categories, creating a significant gap between these datasets and the complex real-world scenarios. Secondly, existing methods follow a two-stage pipeline strategy, which may  lead to the error propagation problem. Additionally, they are difficult to apply in situations where unseen semantic entity categories emerge.
To address the first challenge, we propose a new large-scale human-annotated dataset named {C}omplex {L}ayout form for key information {EX}traction (CLEX), which consists of 5,860 images with 1,162 semantic entity categories. To solve the second challenge, we introduce {P}arallel {P}ointer-based {N}etwork (PPN), an end-to-end model that can be applied in zero-shot and few-shot scenarios. 
PPN leverages the implicit clues between semantic entities to assist extracting, and its parallel extraction mechanism allows it to extract multiple results simultaneously and efficiently.
Experiments on the CLEX dataset demonstrate that PPN outperforms existing state-of-the-art methods while also offering a much faster inference speed. 
\end{quote}
\end{abstract}

\section{Introduction}

\begin{figure*}[t!]%[htbp] % figure1
%备注便于查阅
\centering
\includegraphics[width=15.6cm,clip ]{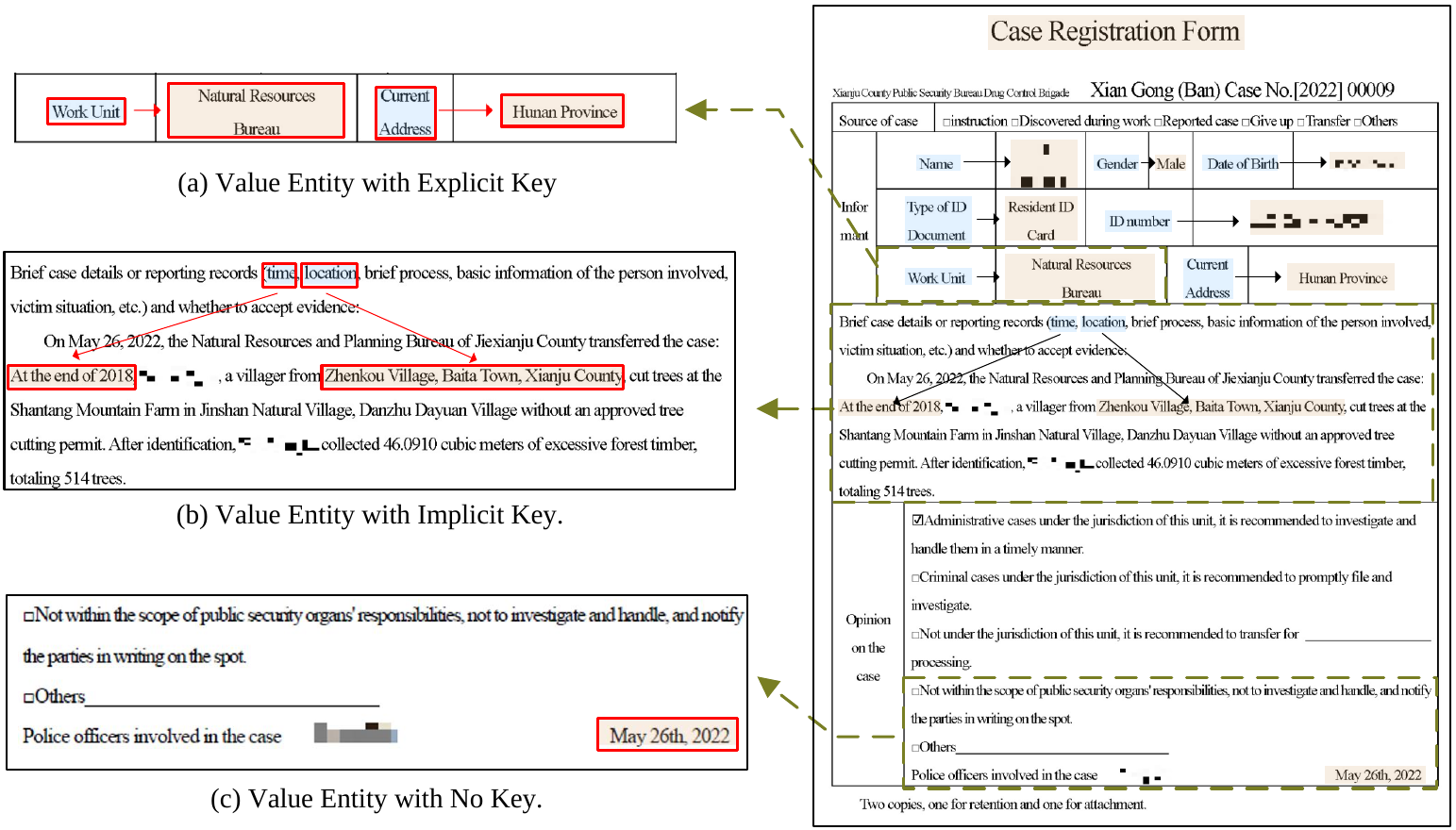}% 可调宽度，自适应高度
\caption{An example of the Key Information Extraction (KIE) task. Existing methods typically require (a) explicit key for extraction, while they are hard to handle (b) implicit key (c) no key situations.
The key and value entities are marked in blue and orange boxes, respectively.
The content of the table is synthesized and some values are decorated for privacy purposes.} %
\label{intro case} %引用标注 
\vspace{-0.6em}
\end{figure*}

Key Information Extraction (KIE) aims to extract value semantic entities from visually rich documents in the semi-structured form \cite{DBLP:conf/mm/CaoLMJGH0L022}, such as forms, digital files, or scanned documents. It typically consists of two sub-steps: entity labeling and entity relation extraction. 
Take Fig.~\ref{intro case}~(a) as an example, an KIE system seeks to first extract the semantic entities (i.e., the text contents with their layout coordinates) \textit{Work Unit} as key and \textit{Natural Resources Bureau} as value, and then predicts key-value relation between the two. Finally identifying \textit{Natural Resources Bureau} as \textit{Work Unit} type. 
% If the format of key is not fixed, the key needs to be further normalized.
KIE has received great attention from the community and can be used in a wide range of downstream tasks, such as business document information registration \cite{DBLP:conf/acl/MajumderPTWZN20}, verification \cite{DBLP:conf/mm/ChengQSH020,  DBLP:conf/icdar/HamdiCJCD21}.

%Existing methods could be roughly categorized into four groups: template-based methods, sequence annotation-based methods, generative-based methods, and linking-based methods.

% 生成式的慢！
% 分为两部分，先讲预训练模型backbone；然后讲序列标注，QA，生成方法
Recently, 
%some datasets and methods have been proposed and applied to the KIE task. 
% Layoutlmv2 \cite{DBLP:conf/acl/XuXL0WWLFZCZZ20}
%For instance, 
several datasets \cite{DBLP:conf/icdar/JaumeET19, DBLP:journals/corr/abs-2104-08836, DBLP:conf/icdar/GuoQLHLD19, DBLP:conf/icdar/HuangCHBKLJ19, park2019cord, DBLP:conf/aaai/WangLJT0ZWWC21} were proposed based on different fields, such as receipts or forms.
%\citet{DBLP:conf/icdar/JaumeET19} proposed FUNSD to extract and structure the textual content of forms. Several other datasets, such as MATEN \cite{DBLP:conf/icdar/GuoQLHLD19}, SROIE \cite{DBLP:conf/icdar/HuangCHBKLJ19}, CORD \cite{park2019cord}, EPHOIE \cite{DBLP:conf/aaai/WangLJT0ZWWC21}, Kleaster NDA and Kleaster Charity \cite{DBLP:conf/icdar/StanislawekGWLK21} were proposed based on different fields.
%and XFUND \cite{DBLP:journals/corr/abs-2104-08836} extend it into 7 different languages.
In addition, a series of pre-trained multi-modal models \cite{DBLP:conf/mm/HuangL0LW22, DBLP:journals/corr/abs-2104-08836, DBLP:conf/emnlp/PengPWLZHCYCZFS22} were presented,  promoting the development of related fields.
%, such as Layoutlmv3 \cite{DBLP:conf/mm/HuangL0LW22}, LayoutXLM \cite{DBLP:journals/corr/abs-2104-08836}, and ERNIE-Layout \cite{DBLP:conf/emnlp/PengPWLZHCYCZFS22}.
Based on those backbones, \citet{DBLP:conf/emnlp/0004ZWCLB21} utilized biaffine model in dependency parsing for KIE. 
\citet{DBLP:journals/corr/abs-2304-07957} and \citet{DBLP:conf/coling/GaoXRXXX22} leveraged the idea of question answering (QA) to extract the value entities. 
After that, \citet{DBLP:conf/eccv/KimHYNPYHYHP22} proposed a generative architecture without Optical Character Recognition (OCR). 
Despite achieving promising results, those datasets and methods still suffer from the following two challenges:

\begin{figure}[t]
	\centering
	\includegraphics[width=7.8cm]{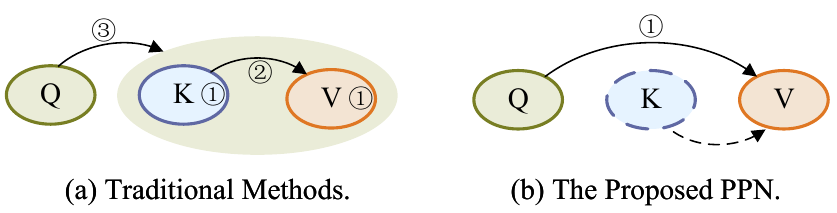}%
	\caption{Comparison between traditional methods and the proposed PPN. Traditional methods have error propagation problem, while PPN  directly leverage question for extracting the value entity with the implicit help of key entity and the key-value relation.  'Q', 'K', and 'V' are short for 'question', 'key', and 'value', respectively. The numbers represent the steps of extraction.
   % needs to first extract the key/value entities and link them based on the key-value relations. Then the keys are normalized to select the key-value pairs that are most relevant to the question as the extraction result. While xxx directly leverage question for extracting the value entity with the implicit help of key entity and the key-value relation.
 } % The results marked with gray background indicate they was excluded.
	\label{intro demons}
	%\vspace{-0.16}
\end{figure}

(1) \textbf{From the task level}: 
%In the practical scenarios, styles of the forms are rich and varied and a large number of unseen semantic entities are emerging. 
Practical scenarios involve a wide variety of form styles and new semantic entities are constantly emerging.
This requires KIE datasets having extensive layout styles and semantic entity categories to simulate such real-world scenario.
%different forms have different layout styles, each of which may contain multiple semantic entity categories. 
However, the layout of existing KIE datasets are relatively fixed and the categories of the semantic entities are not diverse enough. Statistically, the dataset with the most  categories CORD \cite{park2019cord} includes only 30 semantic entity categories. This phenomenon does not align with the need in real-world situations. 
% for new and expanding entity types
%results in a huge disparity between existing datasets and real scenario data. % Moreover, to extract the desired values, 

% they often require explicit key, but in practice there are a large number of implicit key / no key 2) the layout is more fixed, containing few extraction categories.

(2) \textbf{From the method level}: there is an error propagation problem in existing methods. As illustrated in Fig.~\ref{intro demons}~(a), traditional methods need to first extract the key/value entities and link them based on the key-value relations. 
Then the keys are normalized and the key-value pairs that are most relevant to the question is selected as the extraction result.
Any error in any of the steps could result in the extraction errors.
Moreover, existing methods typically require \textit{explicit key} (each value entity could find the corresponding explicit key, as illustrated in Fig.~\ref{intro case}~(a)) to extract the desired value. However, in practice, there is still a large amount of data with an \textit{implicit key} (the key-value relationship is expressed in the text content, as illustrated in Fig.~\ref{intro case}~(b)) or even \textit{no key} (the value entity does not have its corresponding key entity, as illustrated in Fig.~\ref{intro case}~(c)).
%Despite those generative-based methods or QA-based methods could alleviate the problem to some extent, but they are still unsatisfactory in terms of performance and inference speed. 
Although these generation-based or QA-based approaches can alleviate this problem to some extent, they are still unsatisfactory in terms of performance and speed of reasoning. 
Meanwhile, they do not make effectively use of the implicit clues brought from the key entities and key-value pairs, which could intuitively assist in extracting the value entities.

% intuitively, there is implicit relationship between key and value entities, and key can assist in extracting the value. 
%they do not make effectively use of the implicit clues brought from the key entities and key-value paris.
% Additionally, they struggle to be applied to the zero-shot scenario, where new categories are emerging.

To address the first problem, we propose a new human-annotated dataset named \textbf{C}omplex \textbf{L}ayout form for key information \textbf{EX}traction (CLEX), which includes 5,860 form scanned image with 155 form types and 1,162 semantic entity categories. 
%In order to meet the need of the users that they prefer to only provide a small number of annotated samples or even without any samples for training, we mainly consider two situations based on the form types: 
In order to meet the practical needs and make the model adaptable to the scenario of emerging unseen semantic entities, we mainly consider two situations based on the form types: 
(1) \textit{few-shot KIE}, where only a few samples of a certain kind of document layout are used for training, and then tested on samples of the same layout; 
(2) \textit{zero-shot KIE}: directly testing on a certain kind of layout samples without any same type sample for training.

Meanwhile, for solving the second issue, we introduce an end-to-end model named \textbf{P}arallel \textbf{P}ointer-based \textbf{N}etwork for key information extraction (PPN), which simultaneously support the explicit key, implicit key, and no key situations. 
As shown in Fig.~\ref{intro demons}, unlike the traditional approaches, PPN incorporates the implicit clues from the key and key-value relationship while directly extracting the value entities. 
For accelerating the inference speed, PPN formulates KIE as a word-to-word pointer classification problem. In PPN, many questions are concatenated to the input context, and 11 token-linking operations are introduced to compute the outcomes of each question. 
In this manner, many results could be extracted in parallel. 
% In addition, a series of information isolation mechanisms are applied to filter out those impossible results. 
% By training on other known datasets, PPN can transfer the learned knowledge to assist in extracting those value entities in unseen data.

%Through training on the other known datasets, the model can transfer knowledge  to the unseen data.
%not only block the impact between questions and the input context, but also to locate the start and end positions of the input questions.
% key的信息也被融合在了里面

% an end-to-end extraction model is proposed to extract directly to the value while incorporating information from the key. 
%At the same time, pointers can be used to alleviate cross-row and offset problems.

We conduct extensive experiments on the proposed CLEX dataset, the results illustrate that PPN outperforms existing state-of-the-art methods and the inference speed increases 6.4 times compared to traditional QA method. 
In summary, our contributions are listed as follows:

\begin{itemize}
\item We introduce CLEX, a human-annotated complex layout dataset for KIE with 1,162 semantic entities categories, to make the models more adaptable to the real situation.

\item We propose an end-to-end method called PPN, which simultaneously supports explicit key, implicit key, and no key situations. 
PPN parallelly extracts value entities in KIE task while incorporating implicit clues from key entities and key-value pairs.
%It formulates KIE as a word-to-word pointer classification problem, where many semantic entities could be extracted in parallel.

\item The experiment results under both zero-shot KIE and few-shot scenario illustrate that the proposed PPN outperforms state-of-the-art methods with a much faster inference speed. % The relevant dataset and code will be released to facilitate research in related area.
\end{itemize}

%Few-shot KIE and zero-shot experiment results illustrate that the proposed PPN outperforms state-of-the-art methods with a much faster inference speed. 
%This paper considers both few-shot KIE and zero-shot KIE to meet the realistic needs from users. 

% 是否要强调zero-shot/小样本？

% 第二段说现有的方法存在问题： 
% 数据层面：1）往往要求显式key,但是实际中存在大量隐式key/没有key的情况  2）板式较为固定，抽取类别较少 
% 方法上：1）误差传播问题（那个对比图）2）存在部分跨行/偏移问题  
% 在第二页上面画一些例子图

% 第三段说为了解决以上问题，我们提出了:
% 数据层面：我们提出了一个开放类别、开放板式的复杂表单数据集，包含xxx条，xxx类别   
% 方法上：提出了一个端到端的抽取模型，直接抽取到value的同时融入key的信息；同时可以利用指针来缓解跨行、偏移问题

% 第四段介绍实验结果

% contribution:   1)新数据集   2）新的方法   3）我们的方法可以在数据集上取得好的表现，但是仍需要进一步探索

% Based on those backbones,  

\begin{table*}[t!]
\small
\centering
\begin{tabular}{c|ccccc}
\toprule[1pt]
                 & Source                                                                                                    & Image Number & Entity Type & Language     \\ \hline
SROIE            & Receipts                                                                                                  & 1000         & 4           & English          \\
CORD             & Receipts                                                                                                  & 1000         & 30          & English          \\
EPHOIE           & Paper head                                                                                                & 1494         & 10          & Chinese          \\
FUNSD            & Forms                                                                                                     & 199          & 4           & English          \\
XFUND            & Forms                                                                                                     & 199          & 4           & 7 Languages      \\
Kleaster NDA     & EDGAR & 540          & 4           & English         \\
Kleister Charity & UK Charity Commission                                                                                     & 2778         & 8           & English          \\
CLEX (ours)             & Invoices / Certificates                                                                                                          &  5860            & 1162           & Chinese                  \\ 
\bottomrule[1pt]
\end{tabular}
\caption{The comparison between the widely-used KIE datasets and the proposed CLEX dataset. The CLEX dataset is larger in image number and has a wealthier number of entity categories, thus better matching the real situation.}
%'Synth' represents whether the data is synthetic. Please note that the grey tick under column 'Synth' means that only some of the data (320 forms) are fabricated, while most of them (5540 forms) are from real scenes.}
\label{dataset comparison}
\end{table*}

\section{Related Work}
% 相关数据集
% \subsection{KIE Datasets}
\textbf{KIE Datasets. }
There are some datasets are released in the KIE research area. FUNSD \cite{DBLP:conf/icdar/JaumeET19} was proposed to extract and structure the textual content of forms, which contains 199 noisy scanned documents in English with 4 semantic entity categories. After that, XFUND \cite{DBLP:journals/corr/abs-2104-08836} extend it into 7 different languages. Several other datasets, such as MATEN \cite{DBLP:conf/icdar/GuoQLHLD19}, SROIE \cite{DBLP:conf/icdar/HuangCHBKLJ19}, CORD \cite{park2019cord}, EPHOIE \cite{DBLP:conf/aaai/WangLJT0ZWWC21}, Kleaster NDA and Kleaster Charity \cite{DBLP:conf/icdar/StanislawekGWLK21} were proposed based on different fields. Those fields includes scanned tickets, paper heads, receipts, etc. However, the layout of existing KIE datasets are relatively fixed and the categories of the semantic entities are not rich enough, making it hard to be applied to the real scenario. To alleviate this problem, we propose a new dataset named CLEX with 5,860 images and 1,162 semantic entity categories. \\
% 相关FormIE方法
% \subsection{KIE Methods}
% Existing DIE methods could be mainly categorized into three groups: sequence annotation-based methods, question answering-based methods, and generative-based methods.
% Sequence annotation-based methods leverage sequence labeling mechanism for extraction. 
\textbf{KIE Methods. }
Unlike traditional text-only information extraction, key information extraction (KIE) requires further consideration of textual, visual, and layout information. 
%As a result, many pre-trained models such as Layoutlmv3 \cite{DBLP:conf/mm/HuangL0LW22}, LayoutXLM \cite{DBLP:journals/corr/abs-2104-08836}, and ERNIE-Layout \cite{DBLP:conf/emnlp/PengPWLZHCYCZFS22} are emerging.
The KIE task is typically modeled under a pipeline diagram \cite{DBLP:conf/emnlp/0004ZWCLB21}, i.e., first leverage existing document representation models (e.g., Layoutlmv3 \cite{DBLP:conf/mm/HuangL0LW22}, LayoutXLM \cite{DBLP:journals/corr/abs-2104-08836}, and ERNIE-Layout \cite{DBLP:conf/emnlp/PengPWLZHCYCZFS22}) to extract the representations of entities, and then feed these entity representations into different key-value pair extraction models to obtain key-value pairs.
Finally normalize the keys to select the value semantic entity that most relevant to the question as the extraction results.
However, there is an error propagation problem in those methods.
Several methods such as QA-based \cite{DBLP:journals/corr/abs-2304-07957, DBLP:conf/coling/GaoXRXXX22} and generative-based \cite{DBLP:conf/mm/CaoLMJGH0L022} could alleviate this problem, but the performance and inference speed is still far from satisfactory. Additionally, they do not leverage the implicit information brought from the key entities and the key-value pairs.
To handle this problem, we propose PPN model, which parallelly extracts relevant answers based on the word-to-word pointer in the end-to-end manner.

\begin{figure*}[t!]%[htbp] % figure1
%备注便于查阅
\centering
\includegraphics[width=15.6cm,clip ]{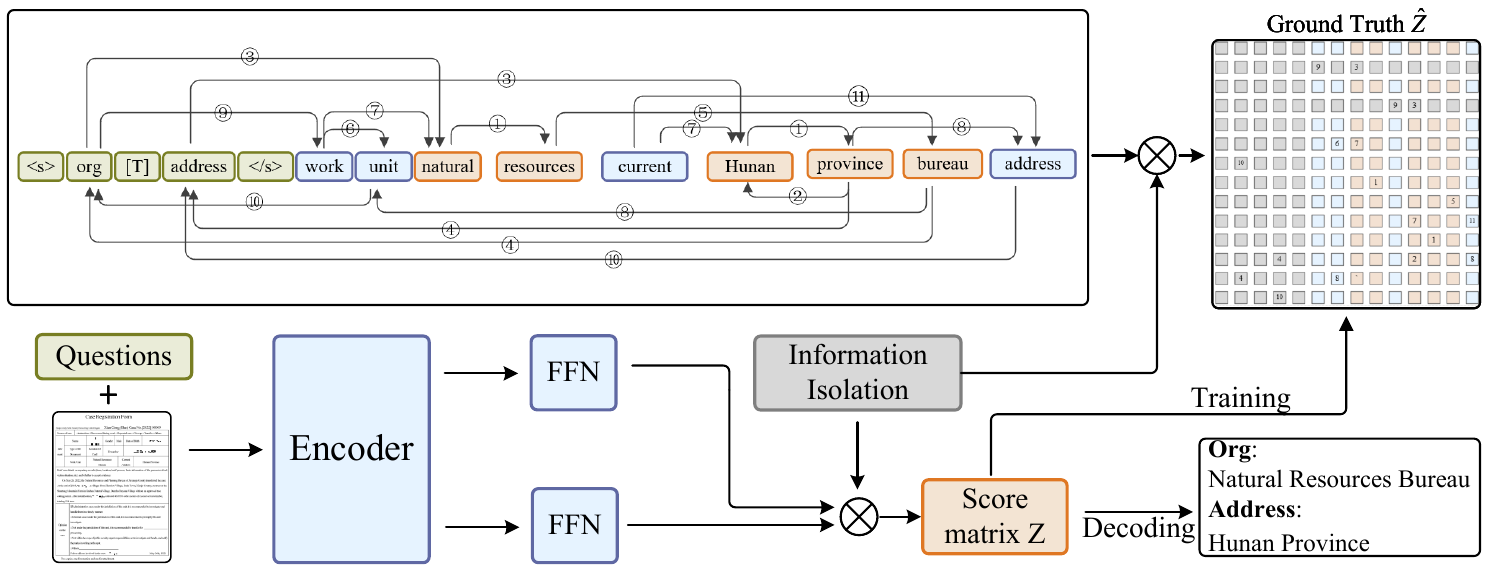}% 可调宽度，自适应高度
\caption{ The architecture of PPN. It mainly has three steps: (1) The questions and context in the form are concatenated and encoded from encoder; (2) Output features are fed to two FFNs, forming the score matrix $Z$ with information isolation mechanism; (3) The score matrix $Z$ are utilized for training and inference. 
} %
\label{model} %引用标注 
\vspace{-0.6em}
\end{figure*} 

\section{CLEX Dataset}
% 花1页左右介绍我们标注的数据集的特点
In this section, we introduce main steps of the data curation process and the statistics of the collected \textbf{C}omplex \textbf{L}ayout form for key information \textbf{EX}traction (CLEX) dataset. \\ \looseness=-1
%\subsection{Dataset Curation}
% 介绍数据集从哪些地方收集的
\textbf{Form Source Selection.}
The sources of CLEX dataset mainly includes the following 4 components: 1) 18 categories of invoice data; 2) 32 categories of card certificates; 3) 86 categories of legal bills; 4) 23 categories of enterprise qualifications, software copyrights and patent certificates. The first category is purchased directly from offline by terminal personnel, while the last three are obtained from search engines through web crawler technology. It is worth noting that each of the total 155 categories of data may include multiple layouts of images. Therefore, CLEX dataset contains more than 155 layouts for training and inference.
\\
% 介绍怎么标注
\textbf{Annotations. }
The data collected in the CLEX dataset may have privacy and copyright issues. 
To solve this problem, we mask out all the content in the original forms and only keep the templates of those forms.
Based on the collected form templates, we combined some content from online public cases of adjudication documents and repopulated these templates with content.
After that, we design the annotation guidelines based on the frequency of occurring categories in each kind of form. 
We invite 20 annotators to label all the data according to our guidelines using a specially designed interface. In the annotation interface, annotators first identify the bounding boxes of key and value entities and assigned a predefined category to the value entity. Then, they annotate the key-value relationship by using drag and drop. Finally, we invite two quality control personnel to randomly inspect 30\% of the samples for each category. If the quality is found to be unsatisfactory, the annotation for that category was revised. The annotation for each category is considered complete when all quality control inspections met the required standards.
\\
% 介绍dataset splits
\textbf{Dataset splits. }
The rules for the division are based on the categories of the different forms. 
In this experiment, we mainly consider zero-shot and few-shot scenarios. In the zero-shot scenario, we randomly divide the training and test sets in the ratio of 7:3, where the form categories in the training and test sets are completely different. Besides, in the few-shot scenario, we move a few data (1/5/10) in the test set to the training set constructed in the zero-shot situation. As a result, a portion of the form data of the same category will be involved in the few-shot training process. 
\\
% \subsection{Dataset Analysis and Statistics}
\textbf{Dataset Analysis and Statistics}
% 数据集分布
After annotation, we got a total of 5,860 images of the forms, which include 1,162 entity types. 
Under the zero-shot situation, we leverage 3,993 forms for training, and the rest 1,867 forms for inference.
Meanwhile, there are 75,438 entities in total, 52,420 with key semantic entities and 23,018 without the key semantic entities.
The comparison of different datasets are illustrated in Table~\ref{dataset comparison}. 
% 画一个类别分布图
% 列一下top20的类别数？

\section{Method}
% In this section, we first present the formulation of KIE, and then introduce how xxx parallelly extract the desired values.
\subsection{Problem Formulation}
% 介绍任务
The input of KIE is a form-like document image $D$ and the context $C$ containing the textual tokens $C=[c_{1}, c_{2}, ..., c_{W}]$ with their corresponding bounding box coordinates. The coordinates of the $i$-th  bounding box could be denoted as $B_i = [x_i^1, y_i^1, x_i^2, y_i^2]$, where $(x_i^1, y_i^1)$ and $(x_i^2, y_i^2)$ are the top-left and bottom-right corner coordinates, and those coordinates could be obtained from the Optical Character Recognition (OCR) tools. 
%The goal of KIE is to extract a list of semantic entities $D=[K_{1}, K_{m}, ..., K_{M};V_{1}, V_{n}, ..., V_{N}]$ and their types $l \in L$, where $K_{m}$ and $V_{n}$ indicate the $m$-th key entity and the $n$-th value entity, respectively. $L$ is the predefined entity label set. 
% Each semantic entity is composed of $k$ words $W_{i} = [w_{i}^{1}, w_{i}^{2}, ..., w_{i}^{k}]$. 
% If there is a directed key-value relation pointing from $K_{m}$ to $V_{n}$, we denote it as $K_{m} \to V_{n}$.
The goal of KIE is to extract a list of semantic value entities $V=[V_{1}, V_{n}, ..., V_{N}]$ and their types $l \in L$, where $V_{n}$ indicates the $n$-th value entity, and $L$ is the predefined entity label set. 
%In this work, we seek to leverage the key entities and key-value relations to assist extracting the value entities. 

%\subsection{PPN model}
%A straightforward approach is to leverage the QA approach for KIE, which could only answer one question at each time, thus heavily reducing the inference speed. To tackle this problem, 
To simultaneously support explicit key, implicit key, no key situations, inspired from \citet{DBLP:conf/aaai/Li00WZTJL22}, PPN formulates KIE as a word-word relation classification task, which could answer many questions in parallel. 
The framework of PPN is illustrated in Fig.~\ref{model}.

\subsection{Encoding}
Following the encoding procedure in \citet{DBLP:journals/corr/abs-2104-08836}, the input embedding consists of two parts: the visual token embedding and the text token embedding. The text input embedding is obtained from the input context $C$. Specifically, PPN first concatenates
many questions [$Q_1$, $Q_2$, ...] to the context $C$, forming the text input $X$: 
\begin{equation}
    X = \left \langle s\right \rangle Q_1[T]Q_2[T]...\left \langle /s \right \rangle C
\end{equation}
where $Q_1$, $Q_2$, ... are different questions. $\left \langle s\right \rangle$ and $\left \langle /s \right \rangle$ are special token from pre-trained models. We leverage a special token $[T]$ to separate the input questions. 
In the experiment, we utilize the likely categories of the semantic entities in the input image $D$ as the questions. 
Since a sample could be concatenated with many questions, to balance the length of the question and the context, we cropped the concatenated question using sliding windows. 

After fed into the pre-trained multi-modal model (LayoutXLM \cite{DBLP:journals/corr/abs-2104-08836} in the experiment), all the tokens in $X$ are serialized into a 1D sequence by reading them in a top-left to bottom-right order, and then mapped into text token embeddings $T$. 
The visual image is first resized into a 224 $\times$ 224 feature map. After a fully connected layer, it is flattened into visual token embeddings $A$.
Then the token and image embeddings are concatenated together. After summing with segment embeddings and position embeddings, the final feature output $H$ is obtained. 
%We leverage the pre-trained multi-modal model (e.g., LayoutXLM \cite{DBLP:journals/corr/abs-2104-08836}) as the encoder to get the representation $H$ of the input $X$. 
%Then Specifically,  Next, it is fed to the encoder to get the corresponding embeddings. 
The encoding process could be formulated as:  
% Then the word sequence is tokenized into a sub-word token sequence,
% \begin{equation}
%     H={\rm Encoder}(X, B, M)
% \end{equation}
\begin{equation}
        H={\rm Encoder}(T, A, B, M)
\end{equation}
where $M$ is the attention mask matrix for isolating the impact of diverse parts, please see section 'Information Isolation' for more details.
% We also conduct experiments on other pre-trained models, please see section xxx for more details. 

After that, to get the relations between different tokens, we first split the hidden states matrix $H$ into $H^i$ and $H^j$, and then feed them into two feedforward neural networks $FFN_{a}$, $FFN_{b}$. We could get the score matrix $Z$:
\begin{equation}
    Z^{i, j} = FFN_{a}(H^{i})^T FFN_{b}(H^{j}) \otimes M^{i, j}
\end{equation}
where $Z^{i, j}$ and $M^{i, j}$ are the score and the mask value from the $i$-th token to the $j$-th token.
Besides, in order to enhance the models' ability to identify entity span, we also add the sinusoidal position embedding \cite{DBLP:conf/nips/VaswaniSPUJGKP17} before calculating score matrix.

% During the training process, we leverage the circle loss \cite{DBLP:conf/cvpr/SunCZZZWW20} to balance the sparse matrices:
% \begin{equation}
%     Z^{j, k} = FFN_{q}(H^{j})^T FFN_{k}(H^{k}) \otimes M^{j, k}
% \end{equation}
% where 

% \subsubsection{WP model}
% 介绍WP模型
%\subsection{Perfixes of xxx}
% 由于单个的拼接很影响速度，因此我们将xxx进行串联

\subsection{Token Linking Operations}
The core of PPN is how to link the head or tail tokens between questions and those semantic entities in the context. 
Take the input sentence in Fig.~\ref{model} as an example, if PPN predicts (org$\to$Natural), (Natural$\to$Resources), (Resources$\to$Bureau), and (Bureau$\to$org), we could get "Natural Resources Bureau" as the "org" type. This operation could be considered as building a directional word graph, which represents the word-word relations. 
The decoding object is to find certain paths from one word to anther word in the graph using the predicted token linking relations.
% “aching in legs” and “aching in shoulders” by NNW relations (aching→in), (in→legs), and (in→ shoulders), and THW relations (legs→aching, Symptom) and (shoulders→aching, Symptom).

\begin{figure}[t]
	\centering  % 7.8
	\includegraphics[width=7.8cm]{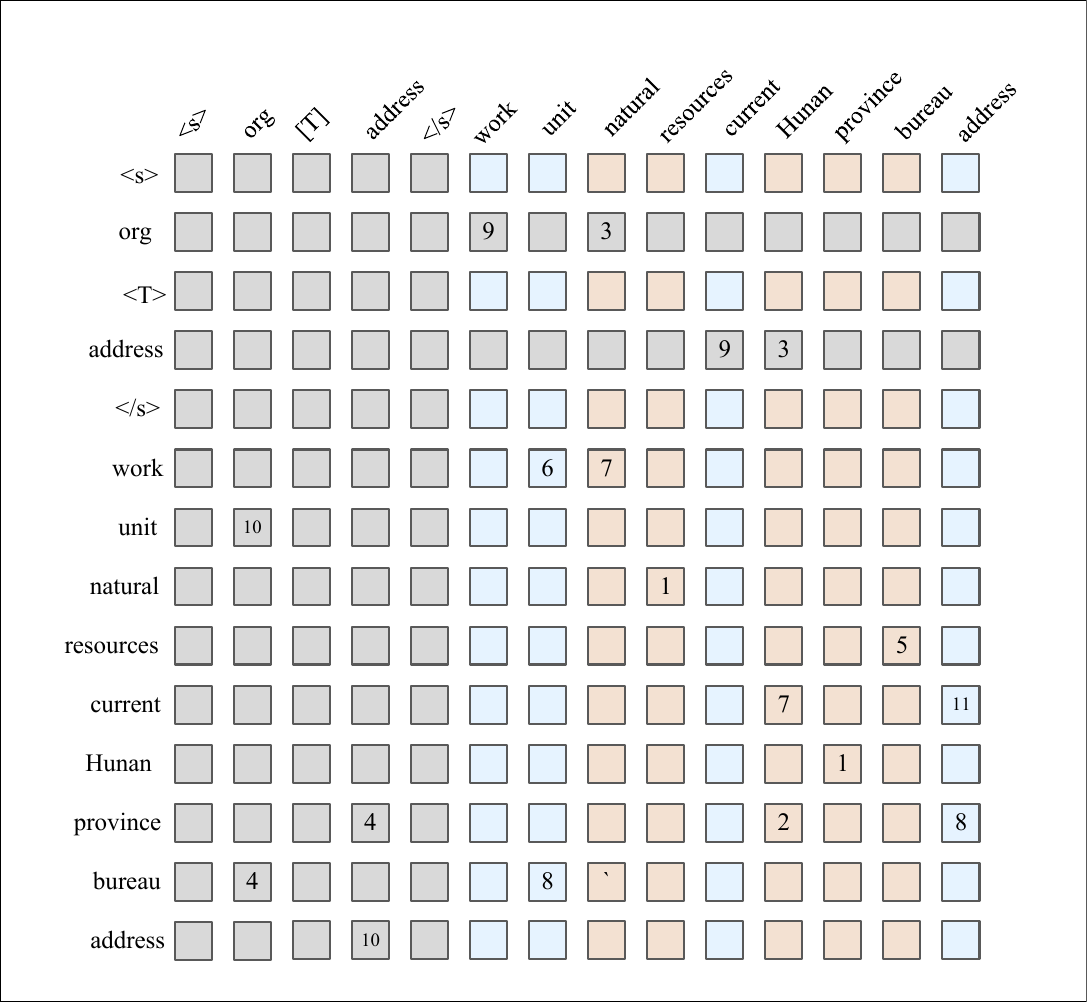}%
	\caption{Illustration of the token linking operations and the information isolation mechanism for building the ground truth matrix $\hat{Z}$. The mask part in isolation operations are shown in gray. 
 %The OCR tool reads horizontally, thus making "Bureau" and "address" are placed on the next line in the sentence.
    } % The results marked with gray background indicate they was excluded.
	\label{mask}
	%\vspace{-0.16}
\end{figure}

To achieve this goal, we conduct five directed token linking operations. Specifically, we design a relation matrix $\hat{Z}$, where different type of linking will be assigned with a different label number.
$\hat{Z}^{i, j}\geq1$ means there is a link between the $i$-th token and the $j$-th token; otherwise, no link exists. We leverage the constructed matrix $\hat{Z}$ as the the ground-truth and let the predicted score matrix $Z$ to fit $\hat{Z}$.
In this section, we expound upon on how to build the 5 types of links with :
\\
\textbf{Value\_Head to Value\_Tail Linking. }
Value head-value tail linking aims at detecting value entity span. Assume the value entity spans from the $i$-th token to the $j$-th token after tokenization, we set $\hat{Z}^{k, k+1}=1$, where $k$ varies from $i$ to $j-1$. For example, the "Hunan province" in Fig.~\ref{model} is the ground-truth value entity, so we let the linking points from "Hunan" to "province" and give this linking a type id as 1. 
\\
\textbf{Value\_Tail to Value\_Head Linking. }
The tail token of each value entity is pointed to the head token of this value entity. 
For instance, given "Hunan province" as the value entity, a connection exists from “province” to “Hunan”. This kind of linking is given a type id as 2. 
\\
\textbf{Question\_Head to Value\_Head Linking. }
The head token of each question is pointed to the head token of the corresponding value entity. 
For instance, the head token\footnote{If a question or semantic entity has only one token, the head and the tail are itself.} "address" of the question will point to the corresponding value entity head token "Hunan". Type id 3 will be given for this kind of linking. 
\\
\textbf{Question\_Tail toValue\_Tail Linking. }
The tail token of each question is pointed to the tail token of the corresponding value entity.
Likewise, the tail of the question "address" will point to the tail token "province".
This kind of linking is given a type id 4.
\\
\textbf{Discontinuous\_Value\_Head to Value\_Tail Linking. }
Since there are problems with value entities such as discontinuities or folded rows, we connect different parts of the same entity first and last. 
As illustrated in Fig.~\ref{mask}, the folded-row problem makes the "natural resources" and "bureau" discontinuous, thus we point from "resources" to "bureau" and give this kind of linking the type id as 5.
If the entity is continuous, we directly point the tail of the entity to the tail itself.

In addition, to further utilize the clues from key entities and the key-value pairs, we offer additional 6 linking operations:
\\
\textbf{Key\_Head to Key\_Tail Linking. }
The head token of each key entity is pointed to its tail token. For example, the head token "work" will be pointed to the tail token "unit" from the key entity "work unit". The linking type id is 6.
\\
\textbf{Key\_Head to Value\_Head Linking. }
The head token of each key entity is pointed to the head token of the corresponding value entity. 
%For instance, the head token of "work unit" will be pointed to "natural". 
The linking type id is 7.
\\
\textbf{Value\_Tail to Key\_Tail Linking. }
The tail token of each value entity is pointed to its corresponding key tail token.
%For example, the tail token "bureau" will be pointed to the tail token of "work unit". 
The linking type id is 8.
\\
\textbf{Question\_Head to Key\_Head Linking. }
The head token of each question is pointed to the head token of the corresponding key entity.
%For example, the question head token "org" will be pointed to the head token of "work unit". 
The linking type id is 9.
\\
\textbf{Key\_Tail to Question\_Tail Linking. }
The tail token of each key entity is pointed to the tail token of the corresponding question.
%For example, the tail token of "work unit" will be pointed to "org". 
The linking type id is 10.
\\
\textbf{Discontinuous\_Key\_Head to Key\_Tail Linking. }
Likewise, the key values could have discontinuous problem, thus we conduct the same operation as in discontinuous value tail-value head linking. For example, given the discontinuous entity "current address", we will connect the "current" and "address". The linking type id is 11.

As for the situation that there is no key entities, we directly skip these 6 key-related linking operations.

% 讲怎么linking 5种
% 进一步介绍怎么同时抽取value，并且融入到模型中
% 添加了5种linking

% \textbf{Token Linking Operations with Keys. }
% \subsubsection{WP+value model}
% 进一步介绍怎么同时抽取value，并且融入到模型中
% 添加了5种linking

\subsection{Information Isolation}
\label{Information Isolation}
% 进一步介绍怎么mask的  3种
In order to filter out those impossible results and further boosting the performance of extraction, we design two mask methods for information isolation in two scenarios:
\\
\textbf{Question Context Isolation. }
In practical, the extracted semantic entities are not likely to appear in the question part. Therefore, to provide this priori knowledge, as illustrated in Fig.~\ref{mask}, we design question-context isolation that mask out the question part during training and inference stages.
\\
\textbf{Question Head/Tail Isolation. }
Since the order of input when we concatenate the questions is self-determined, the start and end positions of each question could be known before input to the model. Therefore, to make it easier to judge the start or end position associated with the questions, we design the question head/tail isolation. With the mask mechanism, more priori knowledge will be given when identifying the linking operation related to the question head/tail.
\\
% \textbf{Question Interactions Isolation. }
% question和context边界检测的mask
% 实体类型边界判断的区域
% 屏蔽标签之间交互的mask

\subsection{Training and inference}
During the training process, we flatten the score matrix $Z$ and the ground truth matrix $\hat{Z}$ into a 1D vector, and leverage the circle loss \cite{DBLP:conf/cvpr/SunCZZZWW20, su2022global} to balance the sparse matrices:
% 这个公式不对
\begin{equation}
\mathcal{L}=\log \left(1+\sum_{\hat{{Z}^{i}} =0} e^{{Z}^i}\right)+\log \left(1+\sum_{\hat{{Z}^{j}} =1} e^{-{Z}^j}\right)
\end{equation}
where we flat the ground truth $\hat{Z}$ and score matrix $Z$, and then calculate the loss function. 

% 简单介绍一下怎么解码，同时怎么训练和inference
During the inference stage, after encoding and obtaining the score matrix $Z$, we could get the  $\tilde{Z}$ by a predefined threshold $\delta$, where $\tilde{Z}$ contains only 0 or 1:  
% 这里不对，这里讲的是训练时构造真值

\begin{equation}
\tilde{Z}^{i, j}=\left\{\begin{array}{cc}
1 & \rm {if} \;\;  Z^{i, j} \geq \delta \\
0 & \rm { otherwise }
\end{array}\right.
\end{equation}

After obtain the directional word graph $\tilde{Z}$, we can utilize it to decode the value entities corresponding to different questions based on the token linking operations.
%The predictions of $\tilde{Z}$ could be considered as a directional word graph, which represents the word-word relations. 
%The decoding object is to find certain paths from one word to anther word in the graph using the predicted token linking relations.
%We can use it to decode the value entities corresponding to different questions. 

\section{Experiment}
\subsection{Experimental Settings}
% 介绍我们用的数据集和评测指标
To verify the capability of the models in the KIE task with complex layouts, we conduct experiments on the proposed CLEX dataset. 
Following previous works \cite{DBLP:conf/icdar/JaumeET19, DBLP:conf/emnlp/0004ZWCLB21}, we take the entity-level precision (P), recall (R), and F1 score as the measure standard. 
%As for the implementation, please refer to Appendix~A for more details.
In the experiment, we leverage the OCR tools to get the text content. We adopt LayoutXLM \cite{DBLP:journals/corr/abs-2104-08836} as the backbone. Please refer to Appendix~A for more details.
% we leverage the OCR tools\footnote{https://duguang.aliyun.com/} to get the text content.
%The models are trained 30 epoch with a 5e-5 learning rate. The models are evaluated once every 500 training steps. We set the batch size and warmup ratio as 8 and 0.1, respectively. The maximum question slice window size is set as 128. The threshold $\delta$ during decoding is set as 0.5. The optimal hyper-parameters are obtained by grid search.
% All the experiments are conducted with pytorch \cite{DBLP:conf/nips/PaszkeGMLBCKLGA19} on one V100 GPU.
% \sufor the main experiment, bsection{Implementation}
% 超参数

\subsection{Experiment Results}

\textbf{Baselines. } Since CLEX is a newly proposed dataset, there is no relevant reports, we select the recent state-of-the-art methods conducted on other datasets as the baselines, including:
(1) \textbf{Sequence Labeling (SL)} \cite{DBLP:journals/corr/abs-2104-08836}, which selects from all the categories and predicts which category each token belongs to; 
(2) \textbf{SimpleDLM} \cite{DBLP:conf/coling/GaoXRXXX22}, which formulates the problem as value retrieval problem; % with arbitrary queries
(3) \textbf{QA} \cite{DBLP:journals/corr/abs-2304-07957}, which models the KIE as a question answering task, where the questions are the likely semantic categories;
%(4) \textbf{UDOP} \cite{DBLP:journals/corr/abs-2212-02623}, which unifies text, image, and layout modalities together.
(4) \textbf{Donut} \cite{DBLP:conf/eccv/KimHYNPYHYHP22}, which is a method based on an OCR-free transformer trained in end-to-end manner.
(5) \textbf{PPN}, which is the proposed pointer-based network that extracts value entities in parrallel; 
(6) \textbf{WPN}, which is a variant of PPN that dose not consider the implicit clues from keys and key-value relations. 
(7) \textbf{ChatGPT}\footnote{https://openai.com/blog/chatgpt}, which has powerful ability in natural language processing tasks. Please see more details of ChatGPT experiment in Appendix~B. 
To evaluate the generation-based methods, we first extract the contents generated for each category, then calculate the start and end positions in the original text, and compare them with the ground-truth positions. Finally we could get the evaluation metrics.  
In order to meet the practical needs and make the model adaptable to the scenario of emerging unseen semantic entities, in the experiment, we mainly consider zero-shot and few-shot scenarios.

The zero-shot experiment results are illustrated in Table~\ref{zero_shot_situation}, we could find that: (1) The SL method perform poorly, which indicates that in the zero-shot scenarios, it is hard to classify from 1,162 categories. Additionally, traditional sequence labeling methods make it difficult to migrate the trained features to unseen format data. %, which affects the model performance. 
(2) Generation-based methods such as ChatGPT and Donut outperforms SL.
%It demonstrates the effectiveness of those generative pre-trained model, 
But they still not performs well, indicating that the generative models have difficulty in the zero-shot KIE task. A possible reason is those generation-based methods are not controllable and generate words that do not exist in the original text.
(3) Compared to the QA method, WPN and PPN have 1.33\% and 2.18\% F1 improvement, respectively. Meanwhile, PPN has reached the state-of-the-art results. It shows that word-word classification problems can effectively model the KIE task with great migration capability. %It also illustrates that those methods have great migration capability.
%compare to the generation-based methods, 

\begin{table}[]
\centering
\begin{tabular}{lccc}
\toprule[1pt]
zero-shot         & P & R & F1 \\ \hline
SL                & 1.62  & 0.28  & 0.48   \\
Donut             & 22.84  & 21.28  & 22.03   \\
ChatGPT           & 56.71  & 54.33  & 55.50   \\
SimpleDLM         & 76.23  & 65.47  & 70.44   \\
QA                & 77.86  & 66.72  & 71.86   \\
%UDOP              &   &   &    \\
WPN                & 78.94  & \textbf{68.22}  & 73.19   \\
PPN               & \textbf{81.62}  & {67.75}  & \textbf{74.04}   \\ 
\bottomrule[1pt]
\end{tabular}
\caption{The zero-shot experiment results.}
\label{zero_shot_situation}
\end{table}

\begin{table}[]
\centering
\begin{tabular}{lccc}
\toprule[1pt]
                  & 1-shot & 5-shot & 10-shot \\ \hline
SL                & 0.23   & 5.13  & 13.94   \\
Donut             & 35.47  & 37.59  & 40.12   \\
SimpleDLM         & 74.05  & 77.86  & 80.79   \\
QA                & 75.33  & 78.79  & 81.39   \\
%UDOP              &   &   &    \\
WPN                & 76.76  & 80.73  & 82.58   \\
PPN               & \textbf{83.96}  & \textbf{80.84}  & \textbf{82.92}   \\ 
\bottomrule[1pt]
\end{tabular}
\caption{The F1 results of few-shot experiments.}
\label{few_shot_situation}
\end{table}

\begin{figure}[t]%[htbp] 
    \centering

	\includegraphics[width=7.0cm]{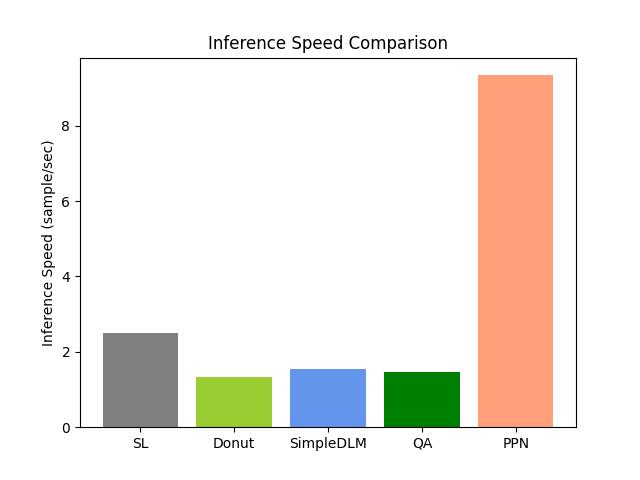}% 可调宽度，自适应高度
	\caption{The speed comparison between different models.}% 图注名称，简单说明
	\label{speed} %引用标注 
	%	\vspace{-0.4em}
\end{figure}

\begin{figure}[t]%[htbp] 
    \centering

	\includegraphics[width=7.8cm]{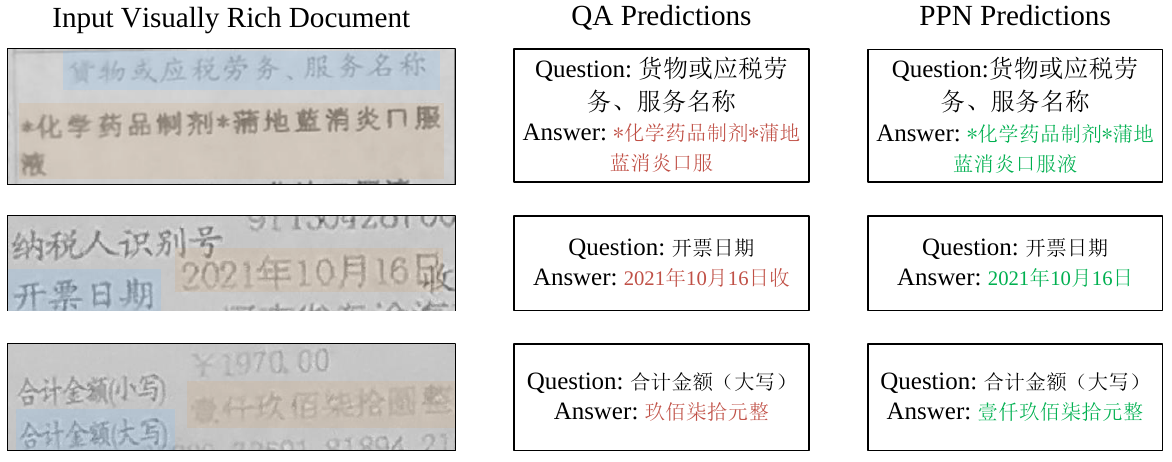}% 可调宽度，自适应高度
	\caption{The case study experiment.}% 图注名称，简单说明
	\label{case_study} %引用标注 
	%	\vspace{-0.4em}
\end{figure}

We also conduct the 1-shot, 5-shot, 10-shot experiments on the CLEX dataset, and the experiment results are shown in Table \ref{few_shot_situation}. 
%, Table \ref{five_shot_situation}, and Table \ref{ten_shot_situation}, respectively. 
We could find that:
(1) As the number of training data in the same category increases (i.e., from 0 to 10), the experimental results of most models improve. This indicates that the models are able to learn the information from a small number of samples and leverage it on the corresponding test set. 
(2) SL still performs poorly in the 1-shot situation, indicating that the model is still unable to learn the migration from a small number of samples. Until be given more samples, 
the performance of SL has a certain improvement.
(3) In all situations, PPN achieves the best experimental results compared to other baseline models. It demonstrates the superiority of the PPN model in few-shot situations. \\
% \subsection{Ablation Study}
\textbf{Ablation Study. } We construct ablation experiments in the zero-shot scenario to demonstrate the validity of the components in PPN, where we eliminate the following modules: 
(1) sinusoidal position embedding (\textit{-sin}); 
(2) the 6 key-related entity linking operations (\textit{-key});
(3) the Question-Context Isolation (\textit{-QCI});
(4) the Question-Head Isolation (\textit{-QHI});
(5) the Question-Tail Isolation (\textit{-QTI});
As illustrated in Table~\ref{ablation}, we could observe that removing the 6 linking operations related to the key entity brings a 0.85\% performance drop. It satisfies our intuition that the key and key-value relations could bring implicit clues to assist extraction. 
Besides, information isolation methods have positive impacts on the final results since bring priori knowledge and block out those false results. Meanwhile, because sinusoidal position embedding can effectively record the relative position relationship between tokens, it brings some performance gains. 
% wp
% 把不带key的拆出来分析一下
% 可以消融position embedding，mask
\begin{table}[]
\centering
\begin{tabular}{lccc}
\bottomrule[1pt]
      & P & R & F1 \\ \hline
PPN   & 81.62  & 67.75  & 74.04   \\
\,\,\,\, \textit{- sin} & 78.82  & 67.55  & 72.75   \\
\,\,\,\, \textit{- key}  & 78.94  & 68.22  & 73.19   \\
\,\,\,\, \textit{- QCI} & 79.85  & 68.92  & 73.98   \\
\,\,\,\, \textit{- QHI} & 79.41  & 68.60  & 73.61   \\
\,\,\,\, \textit{- QTI} & 79.20  & 68.55  & 73.49   \\ 
\bottomrule[1pt]
\end{tabular}
\caption{The ablation study experiment results. }
\label{ablation}
\end{table} \\
% \subsection{Cross-line Study}
% 跨行/偏移单独一个实验吗？问题是现在偏移的数据不多，且全是自己造的数据
% \subsection{Experiment under Full Volume of Data}
\textbf{Experiment under Full Volume of Data. } To explore the maximum capability of the models, we also conduct the experiments under full volume of data. In the experiment, we divide the test set according to the ratio of 7:3.
As illustrated in Table~\ref{full_shot}, we find that: compared to the experiment results in zero-shot and few-shot situations, the performance of all models has further improvement. Meanwhile, the gap between different models is gradually becoming smaller. It shows that the amount of training data has a great influence on the final experimental results. PPN model still has the best performance, showing the superiority of PPN in the full volume data scenario.
% However, the full volume data situation does not fit the real application scenario. The experimental results illustrate the application value of PPN model in the few/zero-shot scenario.
% 探索在全量数据情况下的实验结果。。想验证什么呢？？
\begin{table}[]
\centering
\begin{tabular}{lccc}
\toprule[1pt]
10-shot         & P & R & F1 \\ \hline
SL                & 27.14  & 34.57  & 30.41   \\
Donut             & 50.54  & 50.00  & 50.27   \\
SimpleDLM         & 93.24  & 78.98  & 85.52   \\
QA                & 93.55  & 79.30  & 85.84   \\
% UDOP              &   &   &    \\
WPN                & 93.89  & \textbf{79.58}  & 86.14   \\
PPN               & \textbf{94.57}  & {79.11}  & \textbf{86.15}   \\ 
\bottomrule[1pt]
\end{tabular}
\caption{The results of different methods in the full volume data situation.}
\label{full_shot}
\end{table}  \\
% \subsection{Speed Comparison}
\textbf{Speed Comparison. } We conduce experiment to compare the inference speed between different models. 
We test the models directly on 196 samples of the \textit{VAT invoices (rolled)} category and record the inference time, which includes the time for predicting and decoding.
The experiment results are shown in Fig.~\ref{speed}, we could find that: 
(1) Both SimpleDLM and QA have a relatively low inference speed since they need to ask each sample multiple times, where the number of times is all possible categories for that visually rich document.
(2) The generation-based method such as Dount has a slow inference speed. An important reason is that it is an auto-regressive model, and it needs to generate the answers word by word.
(3) Due to the parallel extraction mechanism, PPN achieves a much faster inference speed. Specially, PPN's inference speed is 6.4 times faster than QA's.
  \\
% \subsection{Case Study}
% 证明我们的可以解决跨行、误差传播等问题
\textbf{Case Study.} In the zero-shot experiment, several typical cases are visualized in Fig.~\ref{case_study}, where we can find that  QA model suffers from errors such as extraction across rows and inaccurate classification of prediction boundaries. But PPN is able to correct complete the KIE task, an important reason is that PPN considers the key and key-value relationship information, thus assisting it to complete the extraction task.

%\subsection{Error Analysis}
% 没预测出来/没照全
% 位置判定不准确
% 

\section{Conclusion}
In this work, in order to simulate the real-world situation that has various types of semantic entities and unseen semantic entities are emerging, we introduce the CLEX, a human-annotated complex layout KIE dataset with 5,860 images and 1,162 semantic entities categories. In addition, we also propose an end-to-end model named PPN, which simultaneously supports explicit key, implicit key, and no key situations. By leveraging 11 token linking operations, PPN incorporates the implicit clues from keys and key-value pairs and extracts value entities in parallel. 
Both zero-shot and few-shot KIE experiments on CLEX dataset illustrate that PPN outperforms state-of-the-art methods with a much faster inference speed.

\clearpage
%\section*{Acknowledgements}
%This research was funded by the National Natural Science Foundation of China (62206267), and the Strategic Priority Research Program of the Chinese Academy of Sciences grant (Y835120378).

% \bibliographystyle{aaai23} 
\bibliography{ref.bib}

\section{Appendices}

\begin{figure*}[h!]%[htbp] % figure1
%备注便于查阅
\centering
\includegraphics[width=15.6cm,clip ]{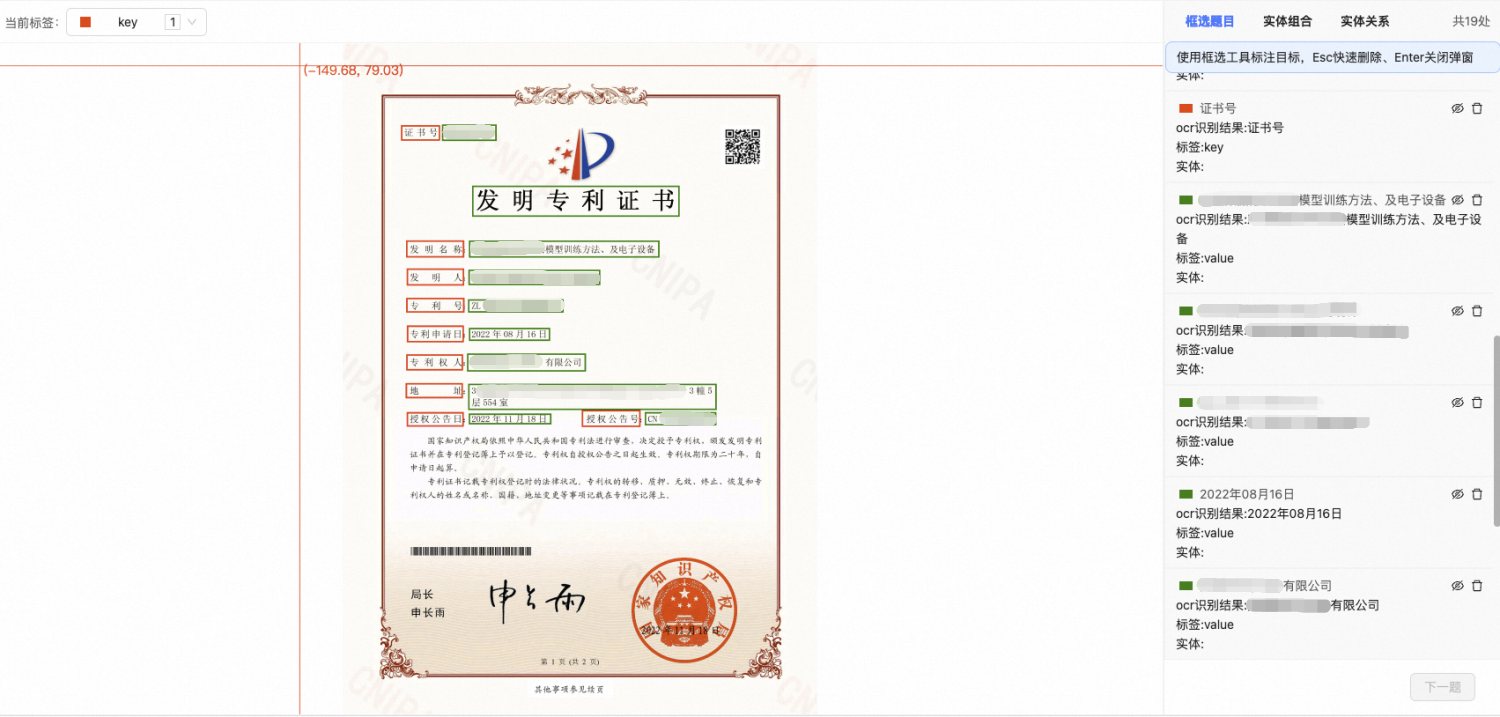}% 可调宽度，自适应高度
\caption{The illustration of the designed annotation interface. In this example, some values are decorated for privacy purposes. } %
\label{interface} %引用标注 
\vspace{-0.6em}
\end{figure*} 

\subsection{Hyper parameter settings}
As for the implementation, we leverage the OCR tools\footnote{https://duguang.aliyun.com/} to get the text content. We adopt LayoutXLM \cite{DBLP:journals/corr/abs-2104-08836} as the backbone. All the experiments are conducted with pytorch \cite{DBLP:conf/nips/PaszkeGMLBCKLGA19} on one V100 GPU. The models are trained 30 epoch with a 5e-5 learning rate. The models are evaluated once every 500 training steps. We set the batch size and warmup ratio as 8 and 0.1, respectively. The maximum question slice window size is set as 128. The threshold $\delta$ during decoding is set as 0.5. The optimal hyper-parameters are obtained by grid search.

\subsection{ChatGPT Experiment Details}
\label{ChatGPT}
We leverage the API from OpenAI to conduct the ChatGPT experiment. We select the \textit{gpt-3.5-turbo-0301} version of ChatGPT for zero-shot inference. The input template is as follows: 

Please extract all entities of category \textit{'[Entity\_type1]', '[Entity\_type2]', ... , '[Entity\_typeN]'} from the text and return the corresponding entity content. The text content is: \textit{[TEXT]}. The return format is: \textit{\{Entity\_type1\}: [Entity1\_content, Entity2\_content, ..., EntityN\_content] or []; \{Entity\_type2\}: [Entity1\_content, Entity2\_content, ..., EntityN\_content] or []; ... ;\{Entity\_typeN\}: [Entity1\_content, Entity2\_content, ..., EntityN\_content] or []}. The return format is json, please answer in Chinese, do not return other content.  \\
where \textit{[Entity\_type1], [Entity\_type2], ..., [Entity\_typeN]} indicate all possible entity categories of this kind of form. \textit{[TEXT]} is the text after OCR recognition. 
The goal is to find the content corresponding to each value semantic entity category.

\subsection{Annotation Interface Illustration}
We specially designed a website for annotation. As illustrated in Fig.~\ref{interface}, the annotators are asked to label the key semantic entities (in the red boxes) and the value semantic entities (in the green boxes). 
They can connect the relevant key-value pairs by using drag and drop, i.e. connecting the corresponding red and green boxes.

\end{document}